\documentclass{article} %
\usepackage{iclr2027_conference,times}

\usepackage{amsmath,amsfonts,bm}

\def\eqref#1{equation~\ref{#1}}

\def\1{\bm{1}}

\DeclareMathAlphabet{\mathsfit}{\encodingdefault}{\sfdefault}{m}{sl}
\SetMathAlphabet{\mathsfit}{bold}{\encodingdefault}{\sfdefault}{bx}{n}

\usepackage{hyperref}
\usepackage{url}

\usepackage{amsmath}
\usepackage{subcaption}
\usepackage{booktabs} 
\usepackage{array}
\usepackage{amssymb}
\usepackage{cleveref}
\usepackage{graphicx}
\usepackage{wrapfig}
\usepackage[table]{xcolor}
\usepackage{arydshln} %
\usepackage{multirow}
\usepackage{amsthm}

\usepackage{float}

\usepackage{hyperref}
\usepackage{fontawesome5}
\newcommand{\github}[1]{\href{#1}{\faGithub\ #1}}

\definecolor{editblue}{RGB}{0, 80, 180}

\newcommand{\method}{\textsc{VFold}}
\newcommand{\methodnoavg}{\textsc{VFold-Naive}}

\title{\method: Symmetry-Aware Cross-Layer Value Cache Compression}

\author{
Neha Verma$^{1}$ \quad
Sungwon Kim$^{1}$ \quad
Kenton Murray$^{2}$ \quad
Kevin Duh$^{1}$
\\
$^{1}$Johns Hopkins University\quad
$^{2}$George Mason University\\
\texttt{nverma7@jhu.edu}, ~~\texttt{kmurra23@gmu.edu},~~\texttt{kevinduh@cs.jhu.edu}
}

\usepackage{todonotes}

\usepackage{booktabs}
\usepackage{array}
\usepackage{adjustbox}        %
\usepackage[table]{xcolor}    %
\usepackage{arydshln}         %

\newcolumntype{R}[2]{%
  >{\adjustbox{angle=#1,lap=\width-(#2)}\bgroup}%
  l%
  <{\egroup}%
}
\newcommand*\rot{\multicolumn{1}{R{45}{1em}}}

\iclrfinalcopy %
\begin{document}

\maketitle

\begin{abstract}
While caching key-value (KV) states accelerates Large Language Model (LLM) decoding, this cache can dominate memory usage at long context lengths.
One solution is to compress this memory by exploiting inter-layer cache similarities. However, most existing techniques necessitate architectural changes to LLMs and incur substantial overhead. 
In this work, we propose a symmetry-aware value cache merging strategy that reduces cache memory while avoiding both harmful performance degradation and architectural overhead during decoding.
Furthermore, we show that this approach can be exploited alongside existing cache compression techniques, composing with high-ratio quantization or key cache pruning to reach compression ratios that
neither method reaches alone, with minimal additional cost.
Ultimately, our findings reveal a major source of underutilized capacity in the value cache, offering a simple yet highly effective direction for scaling context windows under memory constraints.
\end{abstract}

\section{Introduction}
Expanding context windows has unlocked significant new capabilities in LLMs, allowing them to reason over multi-document \citep{bai2024longbench, an-etal-2024-l} or repository-level contexts \citep{jimenez2024swebench, 10.1145/3643757}. %
However, these advancements come at a cost; long contexts shift the bottleneck of autoregressive decoding from compute to memory.
This shift is driven primarily by the presence of the key-value (KV) cache, which stores key and value vectors from attention for prior tokens, in order to prevent their expensive recomputation. 
At sufficiently long contexts, this cache can far surpass the memory requirement of the LLM parameters \citep{MLSYS2023_c4be71ab}. 
For example, for 16GB Llama-3.1-8B (fp16), a batch of 8 inputs at 16K token length also incurs a 16 GB KV cache size, quickly meeting the memory demands of the weights themselves \citep{grattafiori2024llama}. 

Due to this cache memory cost, multiple techniques have been proposed to try to reduce its footprint.
Approaches to KV cache compression can be grouped by the mechanism they employ. 
\textit{Eviction} methods reduce the cache by selecting a subset of paired key/value items to retain while discarding the others \citep{h2o, xiao2024efficient, knorm, kim2026kvzip}. 
\textit{Quantization} methods retain caches but store keys and values at reduced bit width \citep{liu2024kivi, shutova2025cache, zandieh2026turboquant}.
Methods have also exploited low-rank representations of KV caches, and store only rank-reduced representations \citep{wang2025commonkv, chang2026xkv}.
Finally, some methods have looked at sharing caches across layers, whether at initialization or after training \citep{cla, wu2024layer, liu2024minicache, yoco, fusedkv}.

Cross-layer cache sharing, however, has largely been investigated in settings that only allow LLM training from scratch, leaving open the question of how best to apply these strategies to pre-existing models. 
The few methods that do apply to pre-existing models are typically motivated by a common empirical observation: while key and value vectors across layers have low raw pairwise similarity, they are more closely aligned under measures that account for a common subspace \citep{kornblith2019similarity, wang2025commonkv, chang2026xkv}, so a single shared basis can represent several layers' caches with little reconstruction error. 
Existing cross-layer methods frequently realize this shared basis in ways that trade one cost for another. 
Low rank approaches \citep{wang2025commonkv, chang2026xkv} store caches in a compressed low-rank form, forcing ``always-on'' compression, which requires modifications to attention and the reconstruction of vectors during decoding.  
Pre-existing cache merging-based approaches avoid this overhead, but pay in accuracy; MiniCache interpolates caches without re-projection, but degrades heavily on long-context benchmarks \citep{liu2024minicache}. 

\begin{figure*}[t]
     \centering
     \includegraphics[width=\textwidth]{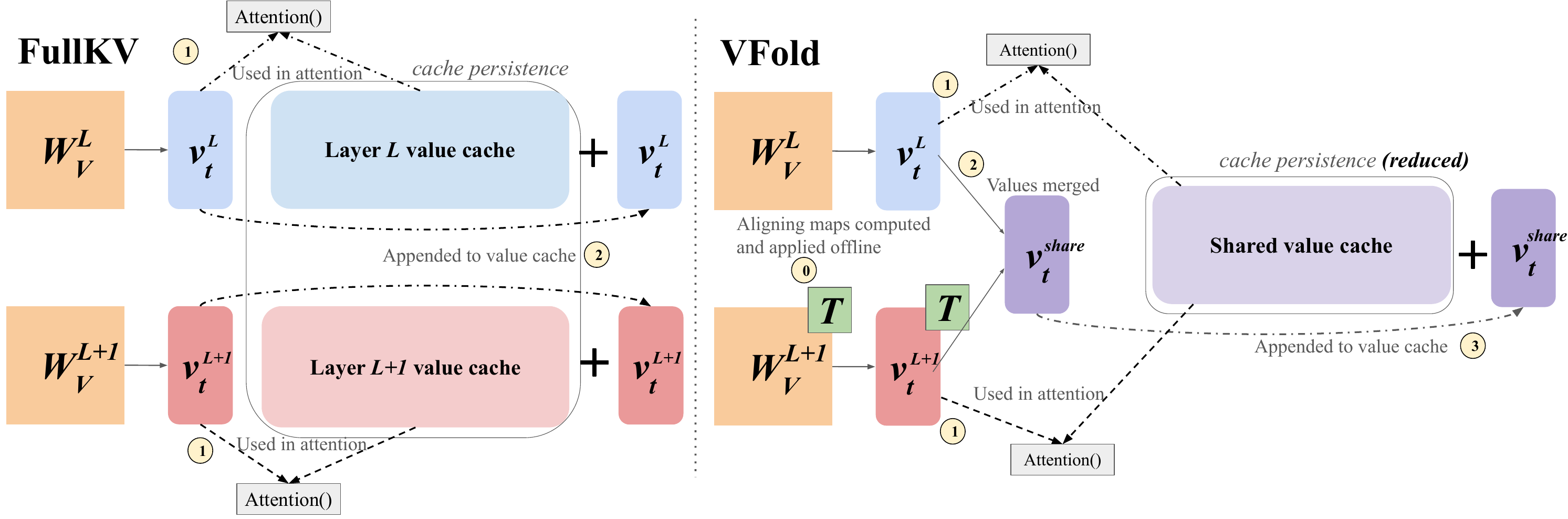}
    \caption{Overview of \method. In the full cache setting (left), individual value caches are stored for layers $L$ and $L+1$. 
    Each layer's new value vector for the $t^\text{th}$ token is appended to its respective cache. 
    In the \method{} setting (right), we first offline fit and absorb a function-preserving linear map $T$ and its inverse into $W_V$ and $W_O$ (not shown), respectively (step 0). The resulting value vectors are used in attention (step 1) unperturbed, and then merged into a single value vector (step 2), that is then appended to a shared inter-layer cache (step 3), resulting in memory savings.
    }
    \label{fig:motivating-figure}
\end{figure*}
To avoid these costs, we propose \method: a simple, query-agnostic method that compresses the KV cache of a pre-trained model by aligning and sharing value caches across adjacent layers (\Cref{fig:motivating-figure}). 
Our contributions are threefold.
\textbf{(1) Analysis:} We demonstrate that although adjacent layers' keys and values show high similarity under Centered Kernel Alignment (CKA) \citep{kornblith2019similarity}, only values are amenable to cross-layer averaging.
\textbf{(2) Method:} We exploit a symmetry within multi-headed attention to fold learned per-head permutations and Canonical Correlation Analysis (CCA) alignment maps directly into attention weights to better align a layer's value cache in terms of a reference layer's value cache basis. 
This model, modified in-place, computes exactly the same function; our method applies without additional fine-tuning, parameters, or changes to attention until cache approximation occurs. 
\textbf{(3) Results:} Halving the value cache via this method (25\% total reduction) retains $>$98\% of full-cache performance on RULER \citep{ruler} and LongBench \citep{bai2024longbench} across three models, outperforming prior cross-layer methods with the lowest decoding overhead and minimal prefill overhead. We also demonstrate that this method composes with 4-bit KIVI and ThinK key pruning for additional compression \citep{liu2024kivi, xu2025think}.\footnote{Code available at \github{https://github.com/nverma1/vfold}}

\section{Related Work}

\paragraph{Cross-layer KV cache compression} Prior work has explored methods that share KV caches across model layers in order to reduce total cache memory consumption. 
Many of these methods create cache sharing patterns at model initialization and pre-train with these patterns, including Cross-Layer Attention \citep{cla}, You Only Cache Once \citep{yoco}, Layer-Condensed KV Cache \citep{wu2024layer}, and FusedKV \citep{fusedkv}. 
However, these methods, as proposed, do not apply to pre-existing models trained with standard per-layer caching.

Some work has indeed investigated the question of reducing cache memory via cross-layer sharing techniques on pre-existing models. 
For example, \citet{liu2024minicache} propose MiniCache, a cross-layer merging technique utilizing spherical interpolation (SLERP) to combine adjacent caches while protecting high importance tokens, but not without a notable cost in accuracy.
\citet{yang2024kvsharer} propose KVSharer, which directly shares dissimilar layers' caches for both prefill and generation.
Other related techniques utilize cross-layer low-rank compression on pre-trained models: xKV \citep{chang2026xkv} compresses after prefill by computing, per request,  a single basis across KV caches for a group of layers. 
CommonKV instead operates on weights by applying a joint SVD across key and value projection matrices of a group, before merging the resulting caches \citep{wang2025commonkv}. 
Finally, recent work has examined using small predictor modules trained on example data to predict key and value vectors based on prior KV vectors across layers. 
AQUA-KV employs this for quantization \citep{shutova2025cache}, and EchoKV employs this for direct compression \citep{ji2026echokv}. 
Our method departs from these predictor methods by avoiding architectural modifications with directly foldable alignment maps.

\begin{figure*}[t]
     \centering
     \begin{subfigure}[b]{0.48\textwidth}
         \centering
         \includegraphics[width=\textwidth]{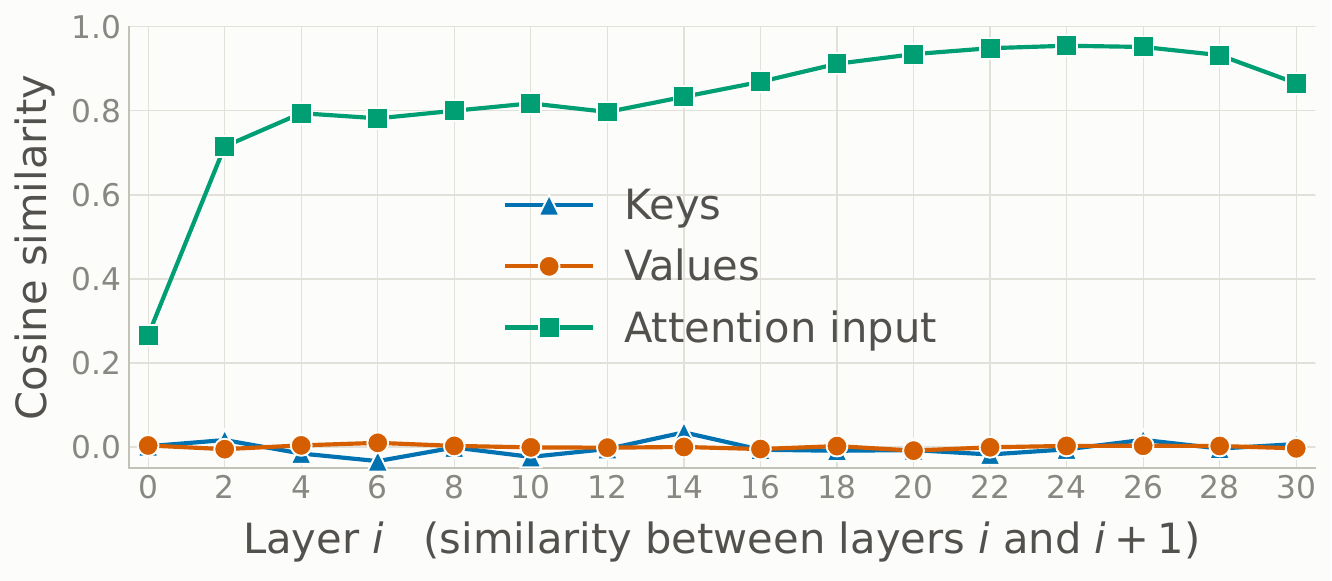}
         \caption{Summary of cosine similarities for adjacent model layers' structures. While inputs to attention layers are quite similar, likely in large part due to residual connections, adjacent layers' key and value caches have nearly zero cosine similarity.}
         \label{fig:cosine_kv}
     \end{subfigure}
     \hfill %
     \begin{subfigure}[b]{0.48\textwidth}
         \centering
         \includegraphics[width=\textwidth]{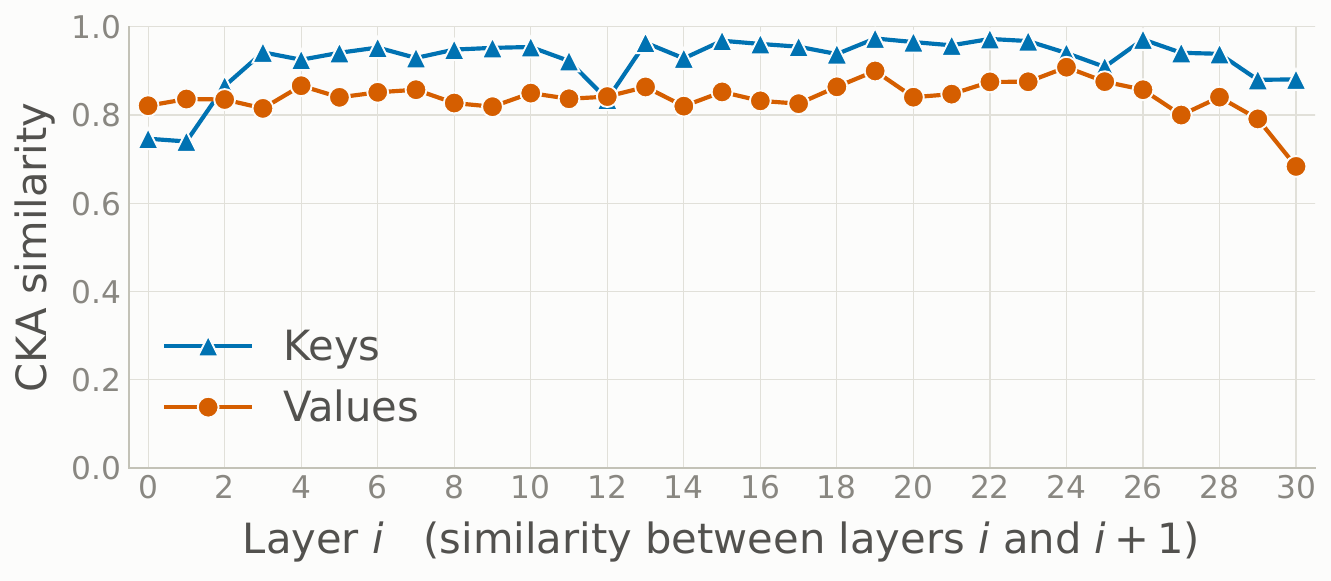}
         \caption{Summary of CKA similarities between adjacent layers' key and value caches. As compared to cosine similarity, CKA similarity is much higher, indicating similarity within latent space rather than direct embedding similarity.}
         \label{fig:cka_kv}
     \end{subfigure}
        \caption{Adjacent KV layer similarities for Llama-3.1-8B-Instruct on GSM8K show near-zero cosine but high CKA similarity.}
    \label{fig:similarities_kv}
\end{figure*}

\paragraph{Symmetry-aware reparameterizations}

Separately, prior work has investigated exploiting weight-space symmetries in Transformers, by folding function-preserving maps into weights in order to achieve various downstream objectives.
For example, SliceGPT folds truncated rotation matrices into LLMs to achieve structured pruning \citep{ashkboos2024slicegpt}; similarly, QuaRot and SpinQuant fold full rotation matrices into LLMs in order to redistribute activation outliers before downstream activation quantization is applied \citep{ashkboos2024quarot, liu2025spinquant}.
RotateKV applies a similar outlier-aware rotation framework specifically to the KV cache \citep{10.24963/ijcai.2025/690}. 
These methods generally focus on orthogonal maps to preserve vector norms.
Extensive work in model merging has exploited various forms of symmetry-aware reparameterizations for aligning several models \citep{ainsworth2023git, pmlr-v235-horoi24a, stoica2024zipit, verma2024merging}.
We adopt similar mapping principles to prepare models for downstream cache compression rather than quantization or model merging objectives.

\section{Preliminary analysis and motivation}
\label{sec:background_motivation}

\paragraph{Inter-layer KV cache similarity} We first corroborate prior work demonstrating that despite having low cosine similarity, adjacent layer caches have high CKA similarity, suggesting a similar underlying basis \citep{kornblith2019similarity}.
CKA (Centered Kernel Alignment) is invariant to rotations and scaling, allowing for similarity measurements that do not depend on an arbitrary basis for comparison.
For this analysis, we use Llama-3.1-8B-Instruct as an example model \citep{grattafiori2024llama}, and GSM8K as an example dataset \citep{gsm8k}.
For each layer, we collect layer inputs, keys, and values on GSM8K, and compute cosine similarities for each adjacent pair, found in \Cref{fig:cosine_kv}. 
Despite having high input similarity, corresponding keys and values across layers do not have notable cosine similarity. 
For CKA similarities, we simply swap out the metric given the same setup for measuring cosine similarity, and report key and value results in \Cref{fig:cka_kv}.\footnote{We use 10,000 tokens for computational feasibility with CKA.}
As seen in the figure, both keys and values have high CKA similarity across layers, corroborating prior results, and motivating cross-layer cache compression.
This finding is similar to those motivating CommonKV and xKV \citep{wang2025commonkv, chang2026xkv}. 
This result suggests that similarity between keys or values between adjacent layers exists, but only up to a change of basis.

\paragraph{Value stability compared to keys}

When the KV cache is compressed in the token dimension, as in eviction, each key/value pair is removed as a unit, so key and value caches are reduced identically.
Non-eviction methods are not bound by the same constraint; they can compress only one of the caches as in ThinK (keys) \citep{xu2025think}, or treat keys and values asymmetrically, as in KIVI \citep{liu2024kivi}, AsymKV \citep{tao-etal-2025-asymkv}, and EchoKV \citep{ji2026echokv}. 
Since we are interested in exploiting cross-layer cache redundancy, we investigate how amenable both keys and values are to simply being averaged across layers.
We again use Llama-3.1-8B-Instruct and GSM8K, and ask how much performance is lost when averaging keys versus when averaging values. 
This means that for each non-overlapping layer pair in the model, two caches are averaged and then stored into memory once, but retrieved by two separate layers for future computations.
For averaging, we do not merge the first 4 tokens and a buffer of the last 128 tokens, as standard in prior literature \citep{xiao2024efficient, liu2024kivi, staniszewski2026kv}, avoiding their inclusion in cache compression to preserve critical performance.
We perform 8-shot Chain-of-Thought completion on GSM8K with greedy decoding \citep{wei2022chain}, and report average accuracy across the averaging and full cache settings in \Cref{tab:kv_sensitivity}. 

\begin{wraptable}{r}{0.38\linewidth}
\vspace{-4mm}
\centering
\caption{GSM8K accuracy when averaging keys vs.\ values across adjacent layers.}
\label{tab:kv_sensitivity}
\begin{tabular}{lr}
\toprule
Setting & Acc. \\
\midrule
Full cache & 80.7\% \\
Avg.\ keys & 2.7\% \\
Avg.\ values & 79.5\% \\
\bottomrule
\end{tabular}
\vspace{-3mm}
\end{wraptable}
From our results, we find values to be far more amenable to merging, despite both keys and values displaying high CKA similarity as seen in \Cref{fig:cka_kv}. 
This asymmetry can potentially be explained by the differing roles keys and values play in attention; while perturbing keys changes how tokens are attended to, which is then amplified by softmax, value perturbation only changes what is retrieved, which then is subject to a linear combination across all prior values.

\paragraph{Attention symmetries and exact value recovery}
\label{sec:test}
Given the observed high inter-layer value cache CKA similarity, we are interested in a low-overhead alignment map to further improve values' mergeability.
To do this, we exploit a natural linear invariance in attention between the value projection ($W_V$) and output projection ($W_O$), also described in prior attention symmetry work \citep{zhang2025beyond}. 

For a given attention head $h$ with inputs $X \in \mathbb{R}^{n \times d_\text{model}}$, values $V^h = XW_V^h$, and attention matrix $A^h$, the head's output is 
\begin{equation}
    \text{Output of Head $h$} = (A^hV^h)W_O^h \in \mathbb{R}^{n \times d_\text{model}}
\end{equation}
which demonstrates the aforementioned linear relationship.
Given this relationship, for any head, we can insert map $T_h \in \text{GL}_{d_h}(\mathbb{R})$\footnote{This is the general linear group of degree $d_h$.} between values and the output matrix:
\begin{equation}
    [A^h(XW_V^h)T_h][T_h^{-1}W_O^h]
    \label{eq:linear_relationship}
\end{equation}
Due to matrix associativity, we can fold $T_h$ into $W_V^h$ and $T_h^{-1}$ into $W_O^h$ entirely offline. Once the transform is applied, the model can run decoding without additional inference or memory cost.
In the case of grouped query attention, we apply $T^{-1}$ to all $W_O$ segments corresponding to the same KV head \citep{ainslie-etal-2023-gqa}.  Critically, this folding technique is in contrast with prior work on low-rank cache compression, where shared SVD low-rank projections drive the aligning of separate layers, at the cost of architectural modification.
These modifications generally result in every new token being compressed.
In our case, full value vectors are recovered after folding, whereas in these low-rank substitutions, exact value vectors are not recoverable due to the ``always-on" nature of these methods \citep{saxena-etal-2024-eigen, palu, wang2025commonkv}.

This same invariance does not apply to the key cache, due to the presence of RoPE in modern architectures \citep{SU2024127063}. 
Because the same linear structure does not apply to $W_K$ with RoPE, any such map would have to commute with the RoPE rotation. This heavily restricts degrees of freedom of the map and conflicts with RoPE being commonly fused into attention kernels for speed \citep{ye2025flashinfer}.

\section{\method{}}
\label{sec:vfold}

Building on the empirical results and symmetry described in \Cref{sec:background_motivation}, we propose our value cache compression method \method.
Our approach consists of offline weight reparameterization followed by runtime cache sharing that enables the cross-layer compression.

\paragraph{Computing alignment maps}
To maximize the similarity of adjacent value caches to prepare them for downstream merging, which \Cref{tab:kv_sensitivity} shows is already promising without alignment, we construct alignment maps using Canonical Correlation Analysis (CCA) on input tokens \citep{pmlr-v235-horoi24a}.
\begin{wrapfigure}{t}{0.5\textwidth}
\vspace{-0.8em}
  \centering
  \includegraphics[width=\linewidth]{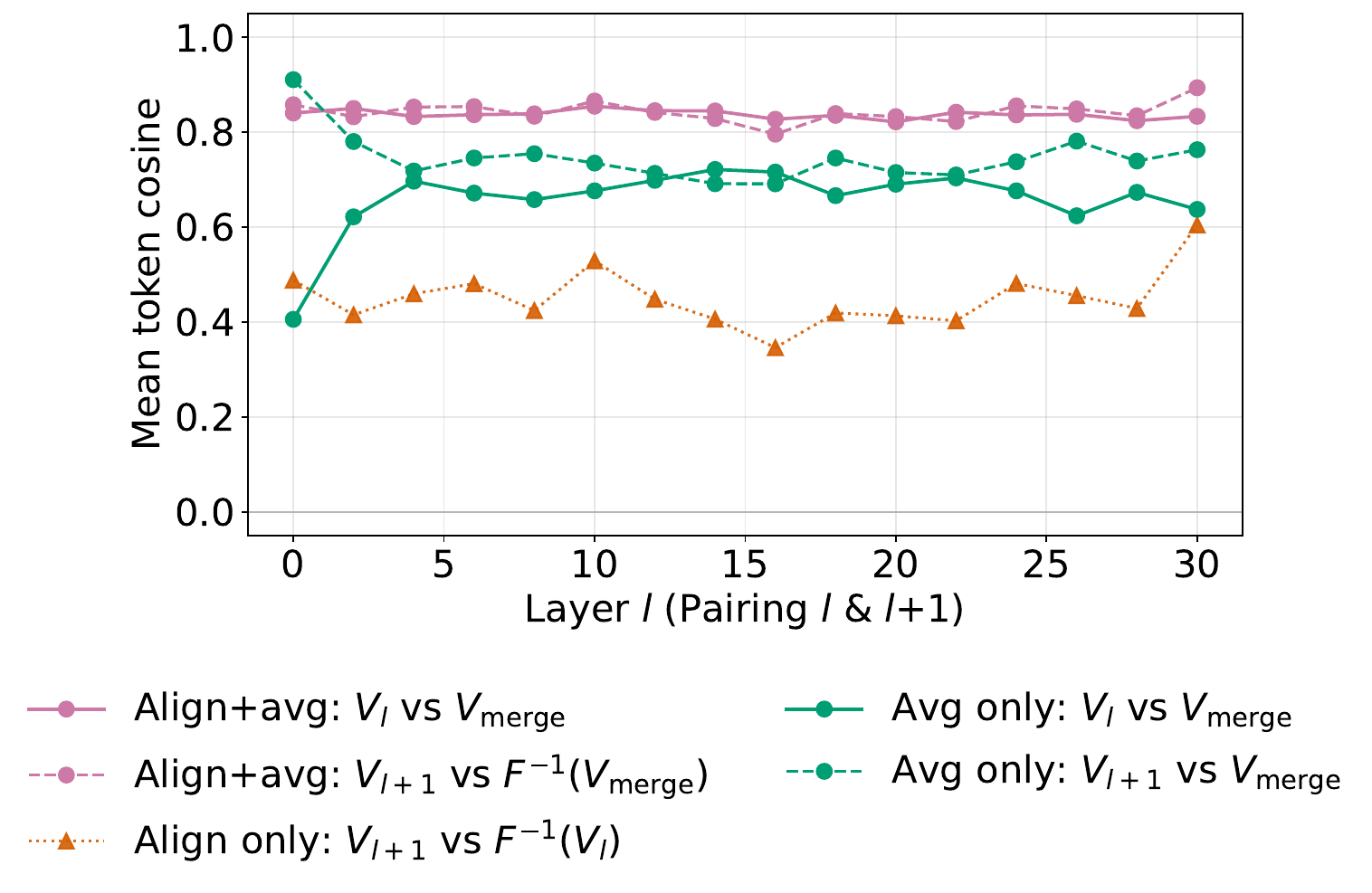}
  \caption{Similarity between combined caches and original value caches. We construct the aligned average as
    $V_{\mathrm{merge}} = \frac{1}{2}\left(V_\ell + F(V_{\ell+1})\right)$
    and compare it with $V_\ell$, for both $F = I$ (unaligned) and CCA. We also map the merged cache back to the coordinate system of layer $\ell+1$ and compare $F^{-1}(V_{\mathrm{merge}})$ with $V_{\ell+1}$. The aligned average achieves cosine similarities close to $0.85$ on held-out data.}
  \label{fig:average}
\end{wrapfigure}
For a base layer $\ell$ and subsequent layer $\ell + 1$, CCA finds projections $U_\ell$ and $U_{\ell +1}$ that project each cache, respectively, into a space with maximum correlation. Rearranging, we define map $T = U_{\ell + 1}U_{\ell}^{-1}$, which is applied to $\ell+1$ alone.

\begin{wrapfigure}{r}{0.35\textwidth}
  \centering
  \vspace{-2em}
  \includegraphics[width=\linewidth]{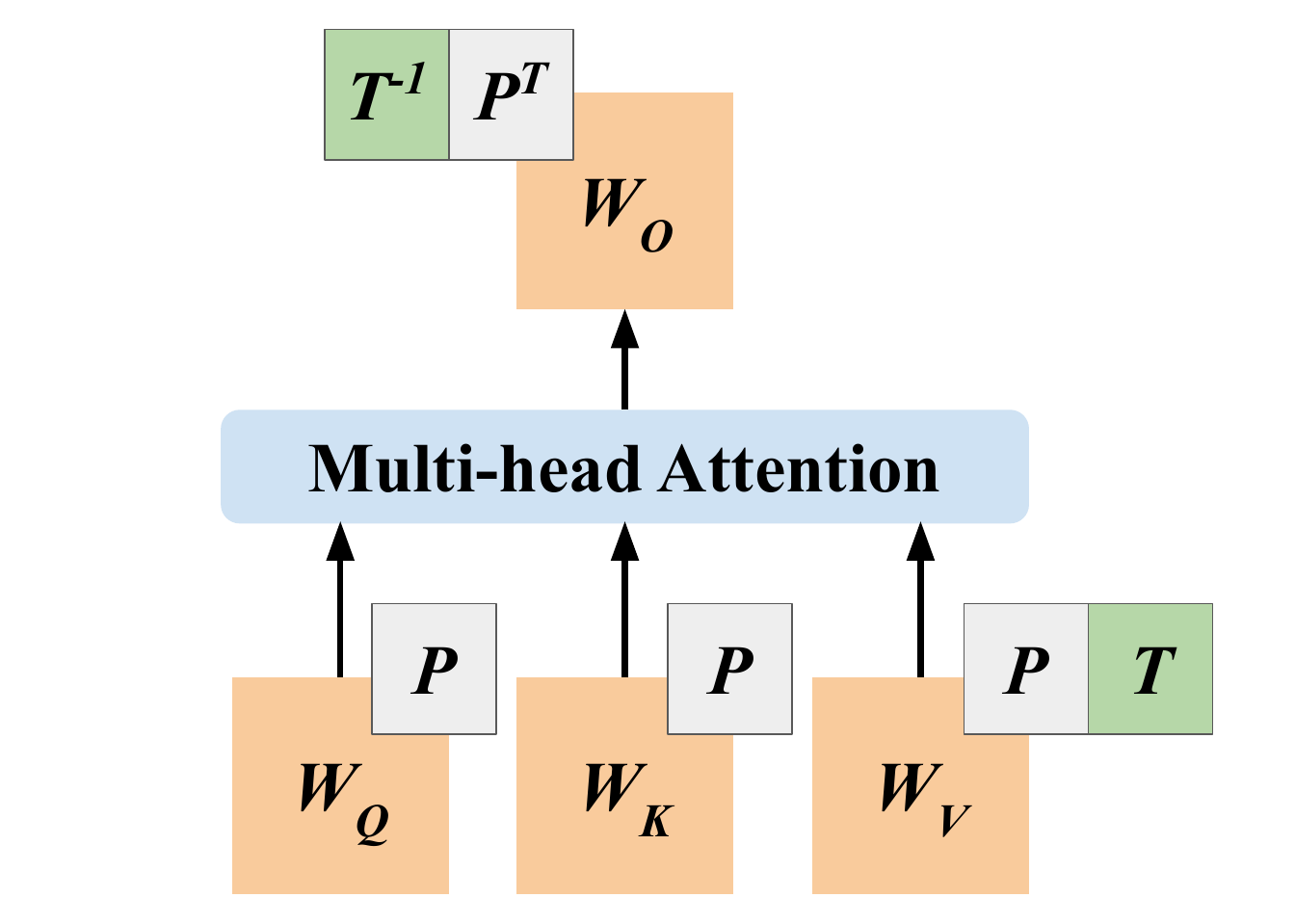}
  \caption{In \method, we apply two main invertible maps to prepare for cache merging. $P$ permutes the ordering of the KV heads, and propagates to all relevant attention weight matrices. $T$, block-diagonally shaped, transforms each head to be more similar to its corresponding head in a reference layer.}
  \label{fig:attn_explainer}
  \vspace{-0.5em}
\end{wrapfigure}

While the above map only applies to a single KV head in multi-head or grouped-query attention, we can extend a map to full value vectors. 
In order to maintain attention invariance, dimensions from different KV heads cannot be mixed; therefore, we optimally pair heads between layers $\ell$ and $\ell+1$ based on their value states.
Then, for every potential head pairing $(i,j)$ between layer $\ell$ and $\ell+1$, we compute CCA map $T_{ij}$, and define the alignment cost via the Frobenius error:
\begin{equation}
C_{ij} = \text{cost}(i,j) = || V_\ell^{h_i}T_{ij}^{-1} - V_{\ell+1}^{h_j}||_F
\end{equation}
We then use the Hungarian algorithm to find the optimal head pairing, or permutation $\pi$, that minimizes the total alignment cost across all $n_{\text{kv}}$ KV heads:
\begin{equation}
\label{eq:perm}
    \pi = \arg\min_{\Pi_{n_\text{kv}}} \sum_{i=1}^{n_\text{kv}} C_{i, \pi(i)}
\end{equation}
Then, defining  $T = \operatorname{blockdiag}(T_1, \dots, T_{n_{\text{kv}}})$, we convert $\pi$ to its matrix form $P$. We apply the maps $P$ to heads of attention weights $W_Q, W_K$ and $W_V$ to reorder them, and then apply $T$ specifically to $W_V$ after heads are re-ordered. More details on these mappings are in \Cref{app:gqa}. Then, to undo this operation, we apply $T^{-1}$ and $P^T$ (applied to heads) to $W_O$.
A summary of this map application appears in \Cref{fig:attn_explainer}.\footnote{For QKNorm, each head uses the same RMSNorm weights, which does not interact with this permutation. RoPE is also applied identically across heads.} More details on CCA and its invertibility are in \Cref{app:cca}.

\paragraph{Aligned cache cosine similarity} To demonstrate the effectiveness of our mapping algorithm, we compute the cosine similarities of aligned and averaged caches with their addends before and after alignment. We compute example caches on Wikitext-2 \citep{merity2017pointer},\footnote{We use 256 batches of length 2048.} and test our maps on 50,000 tokens of Wikitext-2 and held-out GSM8K data. %
The aligned and averaged caches are computed as the following for layers $\ell$ and $\ell+1$, where $F(V) = VPT$
\[
V_{\mathrm{merge}} = \frac{1}{2}\left(V_\ell + F(V_{\ell+1})\right).
\] %
We then measure its similarity to both original caches. For layer $\ell$, we directly compare $V_{\mathrm{merge}}$ with $V_\ell$. For layer $\ell+1$, we map the merged cache back to the original coordinate system and compare $F^{-1}(V_{\mathrm{merge}})$ with $V_{\ell+1}$. As seen in \Cref{fig:average}, the merged cache remains highly similar to both source caches, reaching cosine similarities of approximately $0.85$ on the held-out data. Similar results for in-distribution data are in \Cref{app:cache_sims}.
Note that before, from \Cref{fig:cosine_kv}, adjacent value caches have near-zero cosine similarity without alignment. 

Direct averaging without alignment achieves a cosine similarity of approximately $0.7$, which is higher than the similarity obtained by reconstructing one cache from the other using alignment alone, and close to the theoretical similarity for uncorrelated random vectors.
Nevertheless, aligned merging provides substantially higher similarity to both source caches, compared to either averaging or alignment alone, motivating the use of both in our method.

While we describe \method{} for groups of two layers, we extend this method to larger groups of layers in \Cref{app:more_layers}.
As caches are materialized, layer-by-layer, we combine them, which reduces the realized peak memory during the forward pass.
We temporarily store $V_\ell$ as the merged cache, and then overwrite it with $V_\text{merge}$ after $V_{\ell+1}$ is materialized. We employ a similar running mean for larger group sizes as well. Because the folded model is function-preserving, \method{} only introduces error when a value, after being used in attention, is merged into the shared cache. The current step is always computed with full values, which is not possible in many architecture-modifying methods, like low-rank methods.

\section{Experimental Settings}

\paragraph{Models}
We test our method using 3 models: Llama-3.1-8B-Instruct \citep{grattafiori2024llama}, Mistral-Small-3.1-24B-Instruct \citep{liu2026ministral}, and Qwen3-8B \citep{yang2025qwen3}, containing 32, 40, and 36 layers, respectively. All models use Grouped-Query Attention; Qwen3-8B additionally includes Query-Key Normalization \citep{henry-etal-2020-query}.  
 
\paragraph{Evaluation}
To evaluate capability retention under compression, we use both the LongBench and RULER evaluation benchmarks \citep{bai2024longbench, ruler}. 
LongBench contains real-world long-context tasks covering six task areas: single- and multi-document QA, summarization, few-shot learning, synthetic tasks, and code. 
RULER is a suite of synthetic long context tasks including Needle-in-a-Haystack tasks and aggregative tasks; we use its 16K sequence length version.

\begin{table*}[t]

\centering
\small
\setlength{\tabcolsep}{2.5pt}
\renewcommand{\arraystretch}{1.05}
\caption{RULER (16K) results for KV cache compression methods at 25\% total reduction, matching \method's 50\% value cache reduction. \textbf{Best} average results are bolded, \underline{second-best} are underlined.}
\label{tab:main_ruler}
\resizebox{\textwidth}{!}{
\begin{tabular}{@{}l@{\hspace{4pt}}rrrrrrrrrrrrr@{\hspace{4pt}}r@{}}
\toprule
& \multicolumn{8}{c}{\textbf{Needle In A Haystack (NIAH)}} & \multicolumn{3}{c}{\textbf{Synthetic}} & \multicolumn{2}{c}{\textbf{QA}} & \\
\cmidrule(lr){2-9} \cmidrule(lr){10-12} \cmidrule(lr){13-14}
\textbf{Method} & \textbf{S1} & \textbf{S2} & \textbf{S3} & \textbf{MK1} & \textbf{MK2} & \textbf{MK3} & \textbf{MQ} & \textbf{MV} & \textbf{CWE} & \textbf{FWE} & \textbf{VT} & \textbf{QA1} & \textbf{QA2} & \textbf{Avg} \\
\midrule

\rowcolor{gray!15}[0pt][0pt] \multicolumn{15}{c}{\textit{Llama-3.1-8B-Instruct }} \\

\textit{Full KV} & \textit{100.0} & \textit{100.0} & \textit{100.0} & \textit{99.80} & \textit{100.0} & \textit{99.60} & \textit{99.25} & \textit{98.75} & \textit{81.76} & \textit{91.13} & \textit{98.96} & \textit{79.40} & \textit{53.60} & \textit{92.48} \\

MiniCache & 58.20 & 74.40 & 14.40 & 65.20 & 56.20 & 8.80 & 32.70 & 28.35 & 1.28 & 44.47 & 47.04 & 66.40 & 46.00 & 41.80 \\

MiniCache-V & 96.60 & 92.80 & 84.40 & 84.80 & 63.40 & 9.20 & 75.35 & 80.45 & 17.22 & 63.00 & 49.92 & 66.00 & 47.80 & 63.92 \\

CommonKV & 95.80 & 96.80 & 89.40 & 97.40 & 84.80 & 57.00 & 94.45 & 85.00 & 5.78 & 89.87 & 81.32 & 73.20 & 49.20 & 76.92 \\

\hdashline

\methodnoavg{} & 97.60 & 94.60 & 96.80 & 96.60 & 99.20 & 96.60 & 89.40 & 79.60 & 2.20 & 80.27 & 73.96 & 76.40 & 51.00 & \underline{79.56} \\

\method{} & 100.0 & 100.0 & 100.0 & 99.60 & 100.0 & 98.20 & 97.50 & 96.20 & 75.26 & 88.60 & 98.68 & 79.00 & 51.80 & \textbf{91.14} \\

\midrule

\rowcolor{gray!15}[0pt][0pt] \multicolumn{15}{c}{\textit{Mistral-Small-3.1-24B-Instruct}} \\

\textit{Full KV} & \textit{100.0} & \textit{99.80} & \textit{100.0} & \textit{99.80} & \textit{100.0} & \textit{99.40} & \textit{100.0} & \textit{99.95} & \textit{96.48} & \textit{98.73} & \textit{100.0} & \textit{85.00} & \textit{68.00} & \textit{95.94} \\

MiniCache & 46.40 & 8.60 & 0.00 & 5.60 & 0.80 & 0.00 & 2.95 & 4.45 & 22.86 & 84.80 & 84.96 & 60.00 & 47.00 & 28.34 \\

MiniCache-V & 99.80 & 100.0 & 98.80 & 97.60 & 97.20 & 60.60 & 93.85 & 96.00 & 24.42 & 86.73 & 97.64 & 73.40 & 58.80 & 83.45 \\

CommonKV & 100.0 & 99.80 & 100.0 & 99.80 & 99.80 & 99.20 & 99.90 & 99.95 & 94.02 & 93.87 & 100.0 & 83.00 & 65.00 & \underline{94.95} \\
\hdashline

\methodnoavg{} & 100.0 & 99.80 & 100.0 & 99.80 & 100.0 & 99.40 & 99.75 & 97.90 & 51.32 & 92.53 & 99.24 & 80.80 & 64.20 & 91.13 \\

\method{} & 100.0 & 100.0 & 100.0 & 99.80 & 100.0 & 99.40 & 99.85 & 99.75 & 91.46 & 96.93 & 100.0 & 83.80 & 65.60 & \textbf{95.12} \\

\rowcolor{gray!15}[0pt][0pt] \multicolumn{15}{c}{\textit{Qwen3-8B }} \\

\textit{Full KV} & \textit{100.0} & \textit{100.0} & \textit{99.80} & \textit{99.80} & \textit{100.0} & \textit{99.80} & \textit{99.85} & \textit{99.55} & \textit{83.76} & \textit{93.80} & \textit{100.0} & \textit{75.40} & \textit{59.00} & \textit{93.14} \\

MiniCache & 4.20 & 0.20 & 0.00 & 0.80 & 0.00 & 0.00 & 0.15 & 0.30 & 1.82 & 34.80 & 2.16 & 20.20 & 15.80 & 6.19 \\

MiniCache-V & 97.20 & 95.80 & 94.60 & 95.60 & 67.00 & 19.00 & 73.85 & 79.20 & 29.88 & 83.27 & 91.32 & 61.00 & 49.60 & 72.10 \\

CommonKV & 100.0 & 100.0 & 97.60 & 98.60 & 98.40 & 91.40 & 99.45 & 96.30 & 59.98 & 93.73 & 99.24 & 66.20 & 54.00 & \underline{88.84} \\
\hdashline

\methodnoavg{} & 99.80 & 100.0 & 98.00 & 99.80 & 93.20 & 16.20 & 99.30 & 95.65 & 32.42 & 71.47 & 95.68 & 70.00 & 53.40 & 78.84 \\

\method{} & 100.0 & 100.0 & 99.40 & 99.60 & 99.60 & 98.00 & 99.75 & 99.10 & 73.46 & 92.00 & 99.88 & 74.00 & 56.40 & \textbf{91.63} \\
\bottomrule
\end{tabular}
}
\end{table*}

\paragraph{Baselines}

We compare our method to prior work that also employs cross-layer cache sharing techniques. 
MiniCache is one such method, where adjacent layer key and value caches are combined using spherical interpolation, while keeping important tokens unmerged, and storing norm information separately from cache items \citep{liu2024minicache}. 
By default, caches from layers $N_\text{layers}/2$ to $N_\text{layers}$ are merged, resulting in a 25\% total cache reduction. 
We evaluate both the original \textit{MiniCache} and \textit{MiniCache-V}, which is our modified baseline that applies the same compression across all layers, but exclusively to the value cache to match the same 25\% total reduction. We adopt the default interpolation hyperparameter $t=0.6$, and set the token retention threshold to $\gamma=0.02$.

We additionally include CommonKV, which projects keys and values into a shared low-rank subspace via SVD before cross-layer cache merging \citep{wang2025commonkv}. 
While related to xKV \citep{chang2026xkv}, CommonKV avoids xKV's per-request SVD overhead during prefill, which is costly and less practical for batched requests, by using a single subspace across requests.
Because CommonKV does not natively support QK-Normalization, present in Qwen3-8B, we adapt the method by applying the QK-Norm after reconstructing keys from compressed latents. 
Finally, CommonKV is a prefill-only compression method, but is otherwise closely related to our method. 
We adopt the default CommonKV group size of 4, and fix the number of groups and rank $r$ per layer uniquely per model to match our target compression ratio exactly, remaining close to the default $r=0.7$. Exact hyperparameters per setting are in \Cref{app:commonkv_settings}. 

\paragraph{\method{} Settings}
We evaluate two variants of our method. \method{} includes folded alignment maps and cache averaging. \methodnoavg{} applies only cache averaging.
Comparing these isolates the effect of alignment maps and both demonstrate the utility of post-attention merging. 
For alignment map learning, we calibrate on 128 Wikitext-2 samples with length 2048 tokens.
Following prior work  \citep{xiao2024efficient, jiang2024minference, liu2024kivi, staniszewski2026kv}, we protect 4 attention sinks and a sliding window of 128 recent tokens. Tokens are merged and added to the shared cache as they leave this window. 

\begin{table*}[t]
\centering
\small
\setlength{\tabcolsep}{2.5pt}
\renewcommand{\arraystretch}{1.05}
\caption{LongBench results for KV cache compression methods at 25\% KV reduction, matching \method's 50\% value cache reduction. \textbf{Best} average results are bolded, \underline{second-best} are underlined.}
\label{tab:main_longbench}
\resizebox{\textwidth}{!}{
\begin{tabular}{@{}l@{\hspace{6pt}}rrrrrrrrrrrrrrrr@{\hspace{6pt}}r@{}}
\toprule
\textbf{Methods} & \multicolumn{3}{c}{\textbf{Single-Doc QA}} & \multicolumn{3}{c}{\textbf{Multi-Doc QA}} & \multicolumn{3}{c}{\textbf{Summarization}} & \multicolumn{3}{c}{\textbf{Few-shot}} & \multicolumn{2}{c}{\textbf{Synth.}} & \multicolumn{2}{c}{\textbf{Code}} & \\
\cmidrule(lr){2-4} \cmidrule(lr){5-7} \cmidrule(lr){8-10} \cmidrule(lr){11-13} \cmidrule(lr){14-15} \cmidrule(lr){16-17}
& \rot{\textbf{Nrtv}}
& \rot{\textbf{Qasp}}
& \rot{\textbf{MF-en}}
& \rot{\textbf{Hotpot}}
& \rot{\textbf{2Wiki}}
& \rot{\textbf{Musiq}}
& \rot{\textbf{Gov}}
& \rot{\textbf{QMS}}
& \rot{\textbf{MN}}
& \rot{\textbf{TREC}}
& \rot{\textbf{TQA}}
& \rot{\textbf{SAM}}
& \rot{\textbf{PC}}
& \rot{\textbf{PRe}}
& \rot{\textbf{LCC}}
& \rot{\textbf{RB-P}}
& \textbf{Avg}\\
\midrule
\rowcolor{gray!15} \multicolumn{18}{c}{\textit{Llama-3.1-8B-Instruct }} \\
\textit{Full KV} & \textit{31.3} & \textit{46.8} & \textit{56.6} & \textit{58.1} & \textit{48.9} & \textit{32.5} & \textit{34.5} & \textit{25.3} & \textit{27.0} & \textit{72.5} & \textit{91.7} & \textit{43.7} & \textit{8.6} & \textit{100.0} & \textit{65.1} & \textit{58.3} & \textit{50.0}\\
MiniCache & 20.9 & 31.7 & 40.4 & 40.9 & 24.5 & 20.4 & 23.0 & 21.9 & 22.9 & 66.0 & 79.3 & 36.1 & 11.3 & 86.9 & 47.5 & 46.6 & 38.8\\
MiniCache-V & 26.2 & 40.4 & 48.8 & 49.2 & 36.1 & 27.0 & 21.0 & 22.5 & 24.0 & 61.5 & 91.0 & 41.4 & 3.7 & 75.9 & 44.3 & 42.4 & 41.0\\
CommonKV & 30.3 & 40.9 & 54.5 & 55.4 & 47.8 & 30.2 & 26.0 & 24.1 & 25.0 & 62.0 & 90.0 & 42.9 & 5.1 & 99.5 & 63.7 & 58.3 & 47.3\\
\hdashline
\methodnoavg & 32.5 & 43.1 & 55.6 & 58.3 & 47.0 & 31.8 & 25.1 & 24.5 & 23.3 & 69.5 & 91.3 & 40.4 & 8.6 & 100.0 & 62.7 & 54.6 & \underline{48.0}\\
\method & 32.0 & 45.1 & 57.4 & 57.9 & 48.0 & 31.9 & 31.5 & 25.1 & 25.2 & 72.5 & 91.5 & 43.0 & 7.1 & 99.5 & 65.2 & 57.4 & \textbf{49.4} \\
\rowcolor{gray!15}[0pt][0pt]  \multicolumn{18}{c}{\textit{Mistral-Small-3.1-24B-Instruct }} \\
\textit{Full KV} & \textit{37.5} & \textit{51.5} & \textit{54.8} & \textit{68.9} & \textit{65.7} & \textit{50.2} & \textit{33.1} & \textit{25.0} & \textit{24.7} & \textit{77.0} & \textit{93.2} & \textit{50.3} & \textit{19.5} & \textit{100.0} & \textit{59.3} & \textit{74.4} & \textit{55.3} \\
MiniCache & 31.1 & 38.5 & 41.4 & 54.7 & 50.0 & 36.9 & 28.0 & 23.9 & 23.7 & 67.5 & 87.5 & 41.7 & 23.0 & 82.3 & 32.4 & 50.9 & 44.6\\
MiniCache-V & 32.9 & 42.2 & 53.8 & 64.6 & 56.4 & 42.2 & 19.9 & 23.1 & 21.6 & 77.0 & 93.5 & 45.5 & 4.6 & 100.0 & 52.8 & 66.8 & 49.8 \\
CommonKV & 37.4 & 50.2 & 54.6 & 68.8 & 66.3 & 50.4 & 31.3 & 24.7 & 24.2 & 77.0 & 92.2 & 49.4 & 20.5 & 100.0 & 62.3 & 73.3 & \textbf{55.2} \\
\hdashline
\methodnoavg & 36.6 & 47.6 & 53.5 & 67.7 & 65.7 & 49.6 & 25.2 & 23.6 & 21.1 & 73.5 & 93.8 & 46.1 & 19.5 & 100.0 & 58.1 & 71.7 & 53.3\\
\method & 36.6 & 50.3 & 55.6 & 68.5 & 64.9 & 49.8 & 30.8 & 24.4 & 23.6 & 76.5 & 93.4 & 49.2 & 18.5 & 100.0 & 59.8 & 73.5 & \underline{54.7}\\
\rowcolor{gray!15}[0pt][0pt]  \multicolumn{18}{c}{\textit{Qwen3-8B }} \\
\textit{Full KV} & \textit{27.0} & \textit{48.3} & \textit{53.9} & \textit{59.5} & \textit{43.0} & \textit{36.3} & \textit{33.5} & \textit{24.0} & \textit{24.8} & \textit{71.5} & \textit{90.7} & \textit{44.2} & \textit{1.0} & \textit{100.0} & \textit{69.1} & \textit{65.6} & \textit{49.5} \\
MiniCache & 4.7 & 10.9 & 14.5 & 10.1 & 4.9 & 4.1 & 10.9 & 16.5 & 13.3 & 44.0 & 59.4 & 20.0 & 1.0 & 11.2 & 32.9 & 30.2 & 18.0 \\
MiniCache-V & 19.5 & 32.3 & 42.8 & 40.7 & 26.2 & 23.9 & 24.3 & 21.0 & 21.1 & 70.0 & 90.0 & 36.1 & 0.0 & 95.3 & 59.0 & 50.8 & 40.8 \\
CommonKV & 27.5 & 43.3 & 45.6 & 52.9 & 41.1 & 29.7 & 31.6 & 22.3 & 23.5 & 69.5 & 89.0 & 40.9 & 14.0 & 100.0 & 68.8 & 62.3 & \underline{47.6} \\
\hdashline
\methodnoavg & 25.3 & 42.2 & 47.9 & 58.5 & 41.7 & 35.5 & 25.2 & 22.0 & 20.3 & 67.0 & 90.5 & 40.9 & 2.0 & 99.0 & 67.3 & 62.2 & 46.7 \\
\method & 27.5 & 46.0 & 52.2 & 59.2 & 41.7 & 37.5 & 31.6 & 23.3 & 23.5 & 71.0 & 90.7 & 42.4 & 2.0 & 100.0 & 68.8 & 65.2 & \textbf{48.9} \\
\bottomrule
\end{tabular}
}
\end{table*}

\section{Results}

\subsection{Performance on long-context benchmarks}

Tables \ref{tab:main_ruler} and \ref{tab:main_longbench} report RULER and LongBench results at a 50\% value cache reduction (25\% of the total KV cache). \method{} retains at least 98.3\% of full-cache performance on every model and benchmark. It also achieves the best performance compared to baselines in five of six settings; the exception is Mistral-Small-3.1-24B-Instruct on LongBench, where both CommonKV and \method{} perform similarly to the uncompressed baseline (within 1.2\% of the full cache). 
\method{} outperforms CommonKV more substantially on RULER for Llama-3.1-8B (+14.2pp) and Qwen3-8B (+2.8pp). 

The MiniCache and MiniCache-V results support the key-value asymmetry claims in \Cref{sec:background_motivation}. Applying MiniCache to values (MiniCache-V) substantially improves performance at the same compression ratio. \method{} still outperforms MiniCache-V by a substantial margin on both RULER and LongBench, which demonstrates the importance of design choices in \method, including alignment and compressing the current value vector after attention.

Comparing \method{} with \methodnoavg{} isolates the contribution of the folded alignment maps.
This alignment shows improvements in overall RULER and LongBench scores, with RULER scores improving by over 9 percentage points on average. 
On select tasks, like CWE, alignment recovers a substantial amount of performance lost to direct averaging, like 2.2$\rightarrow$75.3 in the case of Llama-3.1-8B-Instruct. 
This indicates that the alignment is providing substantive value in cache merging,  modeling similarities between caches that can then be exploited during subsequent cache averaging. At greater compression (\Cref{app:more_layers}), alignment recovers even more performance over \methodnoavg. 
However, even without alignment, \methodnoavg{} demonstrates the utility of cache averaging; for Llama-3.1-8B-Instruct, this simplified strategy even surpasses CommonKV on RULER.

\begin{table}[h]
\centering\small
\caption{KV efficiency measurements of Llama-3.1-8B-Instruct at 8192-token prompt with 256 token generation (NVIDIA A100-80G, batch size 1 unless specified, 50\% value reduction). KV memory is measured after generation and includes protected tokens. Max batch is the largest batch that fits a fixed 80 GB memory budget. Timing experiments are averaged over 3 runs.}
\label{tab:efficiency}
\begin{tabular}{lccccc}
\toprule
& \multicolumn{2}{c}{Memory Profile} & \multicolumn{3}{c}{Performance}\\
\cmidrule(lr){2-3} \cmidrule(lr){4-6}
Method & KV red. & KV mem. (GB) $\downarrow$ & TTFT (ms) $\downarrow$ & TPOT (ms) $\downarrow$ & Max batch $\uparrow$ \\
\midrule
\textit{Full KV }& \textit{0\%} & \textit{1.11} &\textit{781} & \textit{21.0}& \textit{33} \\
MiniCache-V & 25\% & 0.84 & 794 & 34.3 & \textbf{38 }\\
CommonKV & 25\% & 0.86 & 885 & 84.7 & 34 \\
\textsc{VFold} & 25\% & 0.84 & \textbf{786} & \textbf{27.8} &\textbf{38} \\
\bottomrule
\end{tabular}
\vspace{-2mm}
\end{table}
\subsection{Efficiency benchmarks}

Many cross-layer KV cache compression methods save decoding memory at the cost of extra computation during prefill, extra per-step computations, or both. 
\Cref{tab:efficiency} reports KV memory, time-to-first-token (TTFT), time-per-output-token (TPOT), and the largest batch that fits an A100-80G GPU budget for Llama-3.1-8B-Instruct at 8K context and 256 generated tokens (see \Cref{app:timing} for additional details).
We measure \method{}, MiniCache-V, CommonKV, and a FullKV baseline.

All three compressed methods achieve close to the theoretical 25\% memory reduction; the small discrepancies are due to protected tokens, and CommonKV's larger discrepancy arises because generated tokens remain uncompressed.
Regarding TTFT, \method{} leaves prefill timing essentially unchanged (781 vs.\ 786\,ms), since its alignment maps are folded into the weights offline, whereas
CommonKV adds 104\,ms (+13\%). 
During decoding, \method{} adds the least overhead of
the three compression methods. MiniCache-V restores merged vectors with their separately-stored norm at each decoding step, and CommonKV expensively reconstructs full keys and values from their low rank stored version at each step.
\method's remaining 6.8\,ms over the full cache likely comes from maintaining the sliding buffer of protected tokens; attention itself runs unmodified. 
The memory savings are also observed in the maximum attainable batch sizes;
\method{} fits 38 sequences versus 33 for the full cache (+15\%), while CommonKV
fits only 34. 
CommonKV's reconstruction step allows for the actual KV memory to remain compressed, but while decoding, all full KV vectors are eventually materialized during the up-projection, resulting in worse memory use.

\subsection{Composition with other compression methods}
\label{sec:composition}

\method{} compresses only the value cache and only along the depth dimension, so it should compose well with methods that compress keys or reduce precision. We test two such compositions.
 
\paragraph{Key cache pruning.} ThinK prunes key channels using query-driven
importance scores~\citep{xu2025think}. 
We prune half of the key channels (25\%
total KV reduction) and add \method{} to reach 50\% cache compression, a ratio that neither key-only nor value-only pruning can reach since it would require removing an entire cache. 
Table~\ref{tab:Think} demonstrates that the two methods pair well to achieve 50\% cache compression, without any detrimental performance drop in their composition. Performance lost when combining the two methods is roughly additive, with the slight exception of Mistral on RULER, indicating that they act largely orthogonally and can compose well for greater compression.

\begin{table}[t]
\centering
\small
\caption{Performance of \method{} and ThinK applied together, as compared to an uncompressed cache baseline, as well as the individual methods.  }
\resizebox{\textwidth}{!}{
\begin{tabular}{lccccccc}
\toprule
& & \multicolumn{2}{c}{Llama-3.1-8B-Inst.} & \multicolumn{2}{c}{Mistral-Small-3.1-24B-Inst.}& \multicolumn{2}{c}{Qwen3-8B} \\
\cmidrule(lr){3-4} \cmidrule(lr){5-6} \cmidrule(lr){7-8}
Method & KV red. & LongBench &  RULER & LongBench & RULER & LongBench & RULER \\
\midrule
Full cache & 0\% & 50.04 & 92.48 & 55.33 & 95.94 & 49.52 & 93.14 \\
\hdashline
ThinK & 25\% & 49.65 & 88.90 & 52.53 & 92.44 & 49.28 & 93.05\\
\method & 25\% & 49.40 & 91.14 & 54.72 & 95.12 & 48.92 & 91.63\\
\hdashline
ThinK+\method  & 50\% & 48.83 & 86.32 & 52.85 & 87.32 & 49.00 & 91.19\\
\bottomrule \\
\end{tabular}
}
\label{tab:Think}
\vspace{-1.5em}
\end{table}

\begin{figure}[t]
  \centering
  \includegraphics[width=0.9\linewidth]{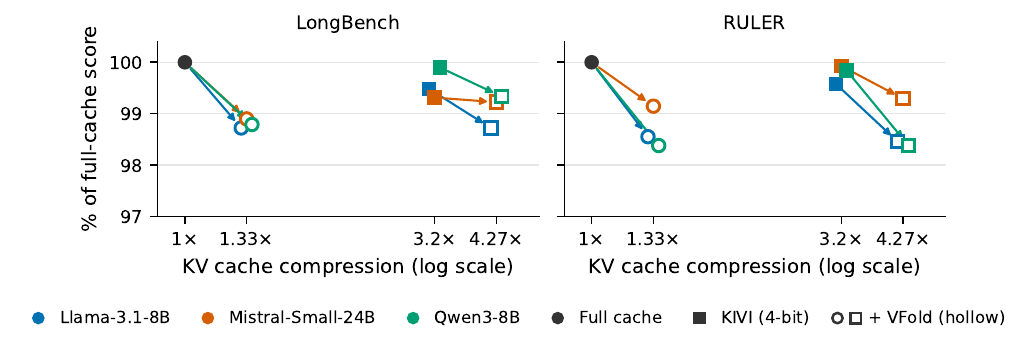}
  \caption{\method{} composed with 4-bit KIVI. Each arrow adds \method{} to a base
  configuration (filled marker) and ends at the combined configuration (hollow
  marker). Every arrow is the cost of adding \method{}. Scores are relative to the full
  cache. Adding \method{} to KIVI is roughly equivalent to adding \method{} alone, allowing for greatly improved compression ratios. }
  \label{fig:kivi}
\end{figure}

\vspace{-0.1em}
\paragraph{Quantization.} KIVI quantizes keys per channel and values per
token~\citep{liu2024kivi}; here, we apply \method{} and then apply 4-bit KIVI. Figure~\ref{fig:kivi} shows that adding \method{} to KIVI improves
compression from $3.2\times$ to $4.27\times$ at almost exactly the performance of \method{} alone: KIVI+\method{} is within 0.3 points of \method{} on every model
and benchmark, and even slightly outperforms it in three of six cases. Quantization and cross-layer merging errors do not compound, consistent with these two methods acting on orthogonal axes. Additionally, since \method{} recovers full value vectors, KIVI applies without modification, which simplifies this composition in practice.

\section{Conclusion}

We introduce \method{}, a low-overhead value cache compression method that exploits weight symmetries in attention to fold low-cost alignment maps into models in order to later combine caches for memory savings and negligible performance degradation. 
\method{} is motivated by several observations, including high inter-layer value cache similarity, high tolerance to cross-layer merging, and an attention-space symmetry unique to the value cache. 
We exploit this symmetry to fold per-head alignment maps into value and output projections to expose cross-layer redundancy while leaving the model's function unchanged; this allows decoding to run on unmodified attention, solely modifying the cache. 
We find that \method{} outperforms similar cross-layer cache reduction methods and an unaligned baseline in both accuracy and efficiency metrics, demonstrating its performance capabilities and the utility of the alignment step. Because \method{} applies to the depth dimension and specifically on the value cache, it composes well with quantization and key cache-only methods.

\bibliography{iclr2027_conference}
\bibliographystyle{iclr2027_conference}

\appendix
\crefalias{section}{appendix}

\section{\method{} map details for Grouped Query Attention}
\label{app:gqa}

All evaluated models in this work include Grouped Query Attention (GQA) with $n_\text{q}$ query heads and $n_{\text{kv}}$ KV heads, of dimension $d_h$ and group size $g = n_\text{q} / n_{\text{kv}}$. 
As for weights, we have $W_Q \in \mathbb{R}^{d_\text{model} \times n_{\text{q}}d_h}$, $W_K, W_V \in \mathbb{R}^{d_\text{model} \times n_{\text{kv}}d_h}$, and $W_O \in \mathbb{R}^{n_\text{q} d_h \times d_\text{model}}$.

The head assignment problem in \Cref{eq:perm} only applies to $n_{\text{kv}}$ heads, since this is how value caches are stored in GQA. 
Let $P_{\pi(i), i} = 1$ so that new KV head $i$
 is original KV head $\pi(i)$.
 Permuting a KV head requires moving all $g$ associated query heads with it, so we expand $P$ via Kronecker product at the appropriate block size:
 \begin{equation}
     P_{\text{kv}} = P \otimes I_{d_h} \qquad P_{\text{q}} = P \otimes I_{g d_h}
 \end{equation}

The Kronecker product allows each block of query heads to move as a unit, in order to maintain their alignment with the corresponding KV heads. Similarly, for $W_O$, $T$ must be expanded to the full model dimension from the GQA dimension. Therefore, we define $T_q = \operatorname{blockdiag}(I_g \otimes T_1, \dots, I_g \otimes T_{n_{\text{kv}}})$, which repeats each KV head's map $T$ for each of its $g$ query heads. 

For the full equations of the attention weights that receive folded maps, we have the following:
\begin{equation}
    W_Q \leftarrow W_Q P_\text{q} \qquad
    W_K \leftarrow W_K P_{\text{kv}} \qquad
    W_V \leftarrow W_V P_{\text{kv}} T \qquad
    W_O \leftarrow T_\text{q}^{-1} P_\text{q}^T W_O
\end{equation}

\section{CCA and its invertibility}
\label{app:cca}

We review details on Canonical Correlation Analysis to describe how we compute the alignment maps.
For each pair of matched KV heads, we find a full-rank $d_h \times d_h$ map using CCA.
For value caches $V_\ell^h, V_{\ell+1}^h \in \mathbb{R}^{n \times d_h}$ over $n$ calibration tokens, we define the following products. For the numerical stability of the inverse square root, we set $\lambda = 10^{-4}$.
\begin{equation}
\Sigma_{11} = \frac{1}{n}(V_{\ell}^h)^T V_\ell^h + \lambda I \qquad
\Sigma_{22} = \frac{1}{n}(V_{\ell+1}^{h})^T V_{\ell+1}^h + \lambda I \qquad \Sigma_{12} = \frac{1}{n}(V_{\ell}^h)^T V_{\ell+1}^h
\end{equation}

We then whiten $\Sigma_{12}$ as follows, and take its SVD:
\begin{equation}
    USV^{T} = \Sigma_{11}^{-1/2}\Sigma_{12}\Sigma_{22}^{-1/2}
\end{equation}
The alignment matrix is the following, after obtaining the SVD:
\begin{equation}
    T = \Sigma_{22}^{-\frac{1}{2}} VU^T \Sigma_{11}^{\frac{1}{2}}
\end{equation}
In order to maintain functional equivalence using the weight symmetry, we must compute the inverse of $T$ for our method. In our implementation, we use the numerical inverse. However, we also note that the inverse of this map is expressible as follows:
\begin{equation}
    T^{-1} = \Sigma_{11}^{-\frac{1}{2}}UV^T\Sigma_{22}^{\frac{1}{2}}
\end{equation}
$T$ is the map from \Cref{sec:vfold} ($U_{\ell+1} U_\ell^{-1}$).

The invertibility of these maps is of interest to determine whether this is a source of error due to numerical stability.
To investigate this, we report the condition number and numerical error of maps used for Llama-3.1-8B-Instruct, for 2-cache groups.
For error, we fold the maps, and then compare (L2 error) the output of the value-output product to the original value-output product. 

\begin{table}[h]
\centering
\caption{Stability of the CCA maps for Llama-3.1-8B-Instruct, over all matched head pairs. $\kappa$ is the condition number. Inverse error compares float32 inverse used when folding with the closed form. Value-output distance is the measure of the distance between $W_V^hW_O^h$ products, with and without mapping.}
\label{tab:cca-stability}
\begin{tabular}{lcc}
\toprule
 & Average & Max \\
\midrule
$\kappa(T)$        & 6.4               & 42.1 \\
Inverse error (fp32)             & $1.4\times10^{-6}$ & $1.6\times10^{-6}$   \\
VO-distance (bf16)              & 0.28\%            & 0.71\%       \\
\bottomrule
\end{tabular}
\end{table}

The results from the table demonstrate that 1) the maps are well-conditioned, so their numerical inverse is stable and close to the exact one, and 2) that the distances between value-output products, before and after folding, are negligible.  

\section{Cache similarities on wikitext}
\label{app:cache_sims}
\begin{wrapfigure}{r}{0.48\textwidth}
  \centering
  \includegraphics[width=\linewidth]{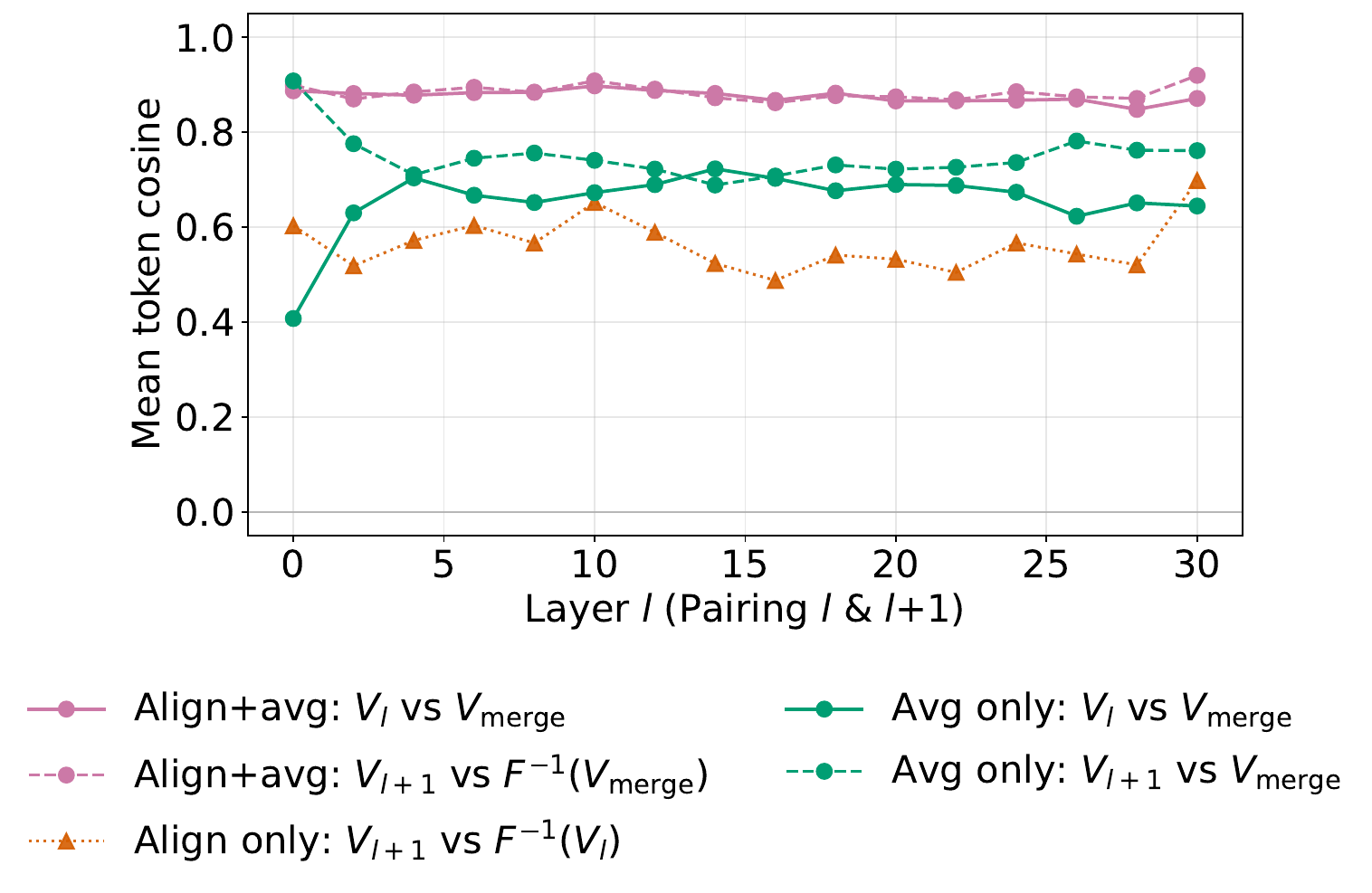}
  \caption{Similarity between combined caches and original value caches. We construct the aligned average as
    $V_{\mathrm{merge}} = \frac{1}{2}\left(V_l + F(V_{\ell+1})\right)$
    and compare it with $V_l$, for both $F = I$ (unaligned) and CCA. We also map the merged cache back to the coordinate system of layer $\ell+1$ and compare $F^{-1}(V_{\mathrm{merge}})$ with $V_{\ell+1}$. The aligned average achieves cosine similarities close to $0.9$ on in-distribution data.}
  \label{fig:wikitext_avg}
  \vspace{-4em}
\end{wrapfigure}
We additionally measure the cache similarities on calibration data for the various value cache representations as described in \Cref{sec:vfold}. 
In \Cref{fig:wikitext_avg}, we report cosine similarities between caches, and find similar patterns to those found in \Cref{fig:average}, with average similarity between aligned caches measuring closer to 0.9 rather than 0.85 as was the case on held-out data.

This suggests that the data used to fit the maps, Wikitext in this case, is general enough to suffice in creating the alignment maps, and/or that the alignment maps are not terribly sensitive to the distribution shift between alignment and test data (GSM8K in \Cref{sec:vfold}).

\section{Extending \method{} to more layers}
\label{app:more_layers}

\paragraph{Extending \method{}} 
The pairwise construction of \method{} ($k=2$) can generalize to groups of
$k$ consecutive layers, which
reduces the value cache to $1/k$ of its original size.
In this section, we discuss this extension, and then results for a $75\%$ value cache
reduction ($37.5\%$ total KV reduction) for $k = 4$. 

Given $k$ and a starting layer $\ell$, we designate a middle layer $\ell + \lfloor k/2 \rfloor$\footnote{Pairs are a special case, where we use $\ell$ rather than $\ell+1$ as the reference.} from the group as our reference, and align every other layer $\ell + j$, $(j=0, \dots, k-1)$ directly to it: we compute head-level CCA costs between
layers $\ell$ and $\ell+j$, solve the head assignment with
the Hungarian algorithm, and fold the resulting permutation
$P_j$ and block-diagonal map $T_j$ into layer $\ell+j$ exactly as in the pairwise
case, with $T_j^{-1}$ and $P_j^{\top}$ folded into $W_O$. Because each map
acts on a different layer's weights, the $k-1$ alignments are independent and the
reparameterized model remains exactly function-preserving. The shared cache
stores
\begin{equation}
\label{eq:k-avg}
  V_{\text{merge}} = \frac{1}{k}\Big(\sum_{j=0}^{k-1} F_j(V_{\ell+j})\Big), \qquad F_j = I \text{ if } j=\lfloor k/2 \rfloor, \text{else } F_j = P_jT_j
\end{equation}
where $F_j$ denotes the alignment folded into layer $\ell+j$. Aligning to a common
reference, rather than chaining pairwise maps, avoids compounding alignment error
across the group. 
Calibration data and protected tokens are identical to the
$k = 2$ setting.

\begin{table*}[h]
\centering
\small
\setlength{\tabcolsep}{2.5pt}
\renewcommand{\arraystretch}{1.05}
\caption{Performance comparisons of KV cache compression methods on LongBench. All results are at 37.5\% reduction, matching the 75\% value cache reduction for \method. We bold the best results, and underline the second-best.}
\label{tab:longbench_375}
\resizebox{\textwidth}{!}{
\begin{tabular}{@{}l@{\hspace{6pt}}rrrrrrrrrrrrrrrr@{\hspace{6pt}}r@{}}
\toprule
\textbf{Methods} & \multicolumn{3}{c}{\textbf{Single-Doc QA}} & \multicolumn{3}{c}{\textbf{Multi-Doc QA}} & \multicolumn{3}{c}{\textbf{Summarization}} & \multicolumn{3}{c}{\textbf{Few-shot}} & \multicolumn{2}{c}{\textbf{Synth.}} & \multicolumn{2}{c}{\textbf{Code}} & \\
\cmidrule(lr){2-4} \cmidrule(lr){5-7} \cmidrule(lr){8-10} \cmidrule(lr){11-13} \cmidrule(lr){14-15} \cmidrule(lr){16-17}
& \rot{\textbf{Nrtv}}
& \rot{\textbf{Qasp}}
& \rot{\textbf{MF-en}}
& \rot{\textbf{Hotpot}}
& \rot{\textbf{2Wiki}}
& \rot{\textbf{Musiq}}
& \rot{\textbf{Gov}}
& \rot{\textbf{QMS}}
& \rot{\textbf{MN}}
& \rot{\textbf{TREC}}
& \rot{\textbf{TQA}}
& \rot{\textbf{SAM}}
& \rot{\textbf{PC}}
& \rot{\textbf{PRe}}
& \rot{\textbf{LCC}}
& \rot{\textbf{RB-P}}
 & \textbf{Avg}\\
\midrule
\rowcolor{gray!15} \multicolumn{18}{c}{\textit{Llama-3.1-8B-Instruct }} \\

\textit{Full KV} & \textit{31.3} & \textit{46.8} & \textit{56.6} & \textit{58.1} & \textit{48.9} & \textit{32.5} & \textit{34.5} & \textit{25.3} & \textit{27.0} & \textit{72.5} & \textit{91.7} & \textit{43.7} & \textit{8.6} & \textit{100.0} & \textit{65.1} & \textit{58.3} & \textit{50.0}\\

MiniCache & 6.3 & 3.7 & 11.2 & 9.5 & 4.1 & 4.0 & 4.4 & 14.1 & 6.3 & 44.0 & 45.9 & 18.2 & 0.9 & 51.4 & 29.2 & 30.6 & 17.7\\

CommonKV & 30.5 & 41.7 & 54.7 & 54.4 & 46.4 & 30.3 & 26.1 & 24.3 & 25.0 & 61.0 & 90.8 & 42.3 & 5.5 & 99.5 & 62.5 & 57.7 & \underline{47.0}\\

\hdashline
\methodnoavg & 30.8 & 30.1 & 44.9 & 55.5 & 44.6 & 29.1 & 18.0 & 21.8 & 18.0 & 36.0 & 91.0 & 38.2 & 9.3 & 100.0 & 57.4 & 49.4 & 42.1\\

\method & 33.1 & 44.1 & 57.1 & 58.3 & 46.2 & 31.1 & 26.8 & 24.7 & 23.5 & 71.5 & 90.4 & 40.9 & 6.9 & 99.5 & 61.3 & 55.2 & \textbf{48.2} \\

\rowcolor{gray!15} \multicolumn{18}{c}{\textit{Mistral-Small-3.1-24B-Instruct }} \\

\textit{Full KV} & \textit{37.5} & \textit{51.5} & \textit{54.8} & \textit{68.9} & \textit{65.7} & \textit{50.2} & \textit{33.1} & \textit{25.0} & \textit{24.7} & \textit{77.0} & \textit{93.2} & \textit{50.3} & \textit{19.5} & \textit{100.0} & \textit{59.3} & \textit{74.4} & \textit{55.3} \\

MiniCache & 14.2 & 25.2 & 27.1 & 40.2 & 26.9 & 20.0 & 11.3 & 19.1 & 13.9 & 68.0 & 84.4 & 31.4 & 3.5 & 55.0 & 25.1 & 38.3 & 31.5\\

CommonKV & 36.1 & 49.5 & 55.2 & 69.5 & 66.1 & 51.1 & 30.7 & 24.5 & 23.9 & 77.5 & 92.5 & 49.0 & 19.5 & 100.0 & 63.4 & 73.2 & \textbf{55.1}  \\
\hdashline

\methodnoavg &32.8 & 39.7 & 46.3 & 65.1 & 63.2 & 46.9 & 20.3 & 21.2 & 15.8 & 41.0 & 93.4 & 44.0 & 17.0 & 100.0 & 54.8 & 64.6 & 47.9\\

\method & 36.9 & 50.0 & 55.4 & 68.4 & 65.5 & 50.2 & 28.5 & 23.7 & 22.1 & 77.0 & 93.6 & 48.6 & 18.0 & 100.0 & 60.3 & 71.9 & \underline{54.4}\\

\rowcolor{gray!15} \multicolumn{18}{c}{\textit{Qwen3-8B}} \\

\textit{Full KV} & \textit{27.0} & \textit{48.3} & \textit{53.9} & \textit{59.5} & \textit{43.0} & \textit{36.3} & \textit{33.5} & \textit{24.0} & \textit{24.8} & \textit{71.5} & \textit{90.7} & \textit{44.2} & \textit{1.0} & \textit{100.0} & \textit{69.1} & \textit{65.6} & \textit{49.5} \\

MiniCache & 0.9 & 4.3 & 5.2 & 2.7 & 2.8 & 1.2 & 4.8 & 13.4 & 7.6 & 30.5 & 21.1 & 9.1 & 0.0 & 0.5 & 30.6 & 24.6 & 10.0\\

CommonKV & 26.7 & 43.6 & 45.5 & 52.9 & 40.7 & 29.0 & 31.9 & 22.6 & 23.6 & 70.5 & 88.1 & 40.0 & 13.0 & 99.5 & 67.9 & 62.1 & \underline{47.3} \\

\hdashline

\methodnoavg & 17.9 & 30.4 & 36.2 & 52.4 & 36.0 & 26.9 & 18.7 & 19.9 & 14.2 & 47.5 & 88.2 & 35.6 & 3.0 & 96.8 & 62.7 & 52.9 & 40.0 \\

\method & 27.8 & 45.5 & 50.5 & 60.1 & 43.1 & 35.7 & 27.7 & 22.7 & 22.0 & 71.0 & 91.2 & 41.3 & 4.0 & 100.0 & 67.8 & 64.0 & \textbf{48.4} \\

\bottomrule
\end{tabular}
}
\end{table*}

\paragraph{Baselines}
At this new compression ratio, we remove MiniCache-V as a baseline because the method uses SLERP to combine caches, and is not specified for groups larger than 2. 
For MiniCache, we apply the method to $N_\text{layers}/4$ until $N_\text{layers}$. For Qwen, we start at layer 8 instead of layer 9 since we need to group layers by 2. 
CommonKV uses the 75\% settings in Table~\ref{tab:rank}.

\paragraph{Merging Execution Schedule}

In order to avoid storing $k$ layers' caches simultaneously during execution before merging, we adopt an incremental averaging approach by combining 2 layers at a time, meaning only one additional cache is ever in buffer during averaging, which reduces peak activation memory during inference.

For the sake of simplicity, we assume all caches are pre-aligned, and write each as $V_i$.
To accommodate the average in \Cref{eq:k-avg}, we write the incremental mean as the following. We utilize relative indexing with 0 representing the first cache in the group. Let $\hat{V}_i$ represent the $i^\text{th}$ buffer. We initialize the buffer with the first cache, $\hat{V}_0 = V_0$.
\begin{equation}
\label{eq:running_avg}
     \hat{V}_i = \alpha_i \hat{V}_{i-1} + (1-\alpha_i)V_i, \qquad \alpha_i = \frac{i}{i+1}
\end{equation}

Briefly, we demonstrate this incremental mean gives the average from \Cref{eq:k-avg}. 

\textit{Claim}. Given \Cref{eq:running_avg}, $\hat{V}_k = \frac{1}{k+1} \sum_{i=0}^{k} V_i$ for all $k$.
\begin{proof}
    We prove the claim by induction. The base case holds trivially, $\hat{V}_0 = V_0$. 

    Now, assume for induction that $\hat{V}_{k-1} = \frac{1}{k} \sum_{i=0}^{k-1} V_i$. Then,
    \begin{equation}
        \hat{V}_{k} = \frac{k}{k+1} \cdot \frac{1}{k} \sum_{i=0}^{k-1}V_i + \frac{1}{k+1}V_k = \frac{1}{k+1}\sum_{i=0}^k V_i
    \end{equation}
\end{proof}

\paragraph{Results} Table~\ref{tab:longbench_375} reports LongBench results for $k = 4$, and Table~\ref{tab:ruler_375} reports RULER. On LongBench, \method{} retains 96.4\%, 98.4\%, and 97.8\% of full-cache performance on Llama, Mistral, and Qwen, outperforming CommonKV on Llama (+$1.2$) and Qwen (+$1.1$) and trailing it slightly on Mistral (-$0.7$). 
RULER demonstrates greater performance spread at this ratio; \method{}  outperforms CommonKV by a wide margin on Llama (83.3 vs.\ 63.7), where CommonKV fails on the hardest multi-key task (MK3: 4.0 vs.\ 89.4). CommonKV is ahead on Mistral ($-0.6$), mostly due to CWE, and on Qwen ($-2.3$), where the gap comes almost entirely from MK3 (41.2 vs.\ 74.2) and FWE. 
On the remaining eleven tasks, \method{} is within 0.4 points of CommonKV or ahead.
 
 The value of alignment grows markedly with group size. 
On LongBench, the gap between \method{} and \methodnoavg{}{} widens from $\approx 2$ points at $k = 2$ to $\approx7$ points at $k = 4$. On RULER this discrepancy widens further, ranging from between 4.0--12.8 points at $k=2$ to between 37.2--67.1 points at $k=4$.
Unaligned averaging (\methodnoavg{}) of four caches collapses on RULER (17.3, 56.9, and 18.1 on Llama, Mistral, and Qwen), while aligned averaging does not. 
MiniCache
degrades severely at this ratio on both benchmarks.

\begin{table*}[t]
\centering
\small
\setlength{\tabcolsep}{2.5pt}
\renewcommand{\arraystretch}{1.05}
\caption{Performance comparisons of KV cache compression methods on RULER, at 16K length. All results are at 37.5\% reduction, matching the 75\% value cache reduction for \method. We bold the best results, and underline the second-best.}
\label{tab:ruler_375}
\resizebox{\textwidth}{!}{
\begin{tabular}{@{}l@{\hspace{4pt}}rrrrrrrrrrrrr@{\hspace{4pt}}r@{}}
\toprule
& \multicolumn{8}{c}{\textbf{Needle In A Haystack (NIAH)}} & \multicolumn{3}{c}{\textbf{Synthetic}} & \multicolumn{2}{c}{\textbf{QA}} & \\
\cmidrule(lr){2-9} \cmidrule(lr){10-12} \cmidrule(lr){13-14}
\textbf{Method} & \textbf{S1} & \textbf{S2} & \textbf{S3} & \textbf{MK1} & \textbf{MK2} & \textbf{MK3} & \textbf{MQ} & \textbf{MV} & \textbf{CWE} & \textbf{FWE} & \textbf{VT} & \textbf{QA1} & \textbf{QA2} & \textbf{Avg} \\
\midrule

\midrule
\rowcolor{gray!15}[0pt][0pt] \multicolumn{15}{c}{\textit{Llama-3.1-8B-Instruct }} \\

\textit{Full KV} & \textit{100.0} & \textit{100.0} & \textit{100.0} & \textit{99.80} & \textit{100.0} & \textit{99.60} & \textit{99.25} & \textit{98.75} & \textit{81.76} & \textit{91.13} & \textit{98.96} & \textit{79.40} & \textit{53.60} & \textit{92.48} \\

MiniCache & 3.60 & 0.40 & 0.00 & 0.40 & 0.20 & 0.00 & 0.00 & 0.00 & 0.10 & 8.00 & 6.00 & 21.20 & 15.80 & 4.28\\

CommonKV & 96.80 & 73.60 & 63.00 & 83.40 & 62.00 & 4.00 & 76.55 & 62.65 & 14.48 & 88.80 & 80.04 & 73.40 & 49.60 & 63.72\\

\hdashline

\methodnoavg{} & 0.20 & 19.60 & 16.00 & 19.40 & 30.00 & 0.40 & 3.35 & 2.65 & 0.04 & 3.87 & 16.44 & 67.20 & 45.60 & 17.29\\

\method{} & 99.20 & 90.80 & 92.80 & 94.00 & 94.40 & 89.40 & 93.55 & 84.35 & 41.20 & 84.07 & 91.12 & 77.00 & 50.80 & 83.28 \\

\midrule

\rowcolor{gray!15}[0pt][0pt] \multicolumn{15}{c}{\textit{Mistral-Small-3.1-24B-Instruct}} \\

\textit{Full KV} & \textit{100.0} & \textit{99.80} & \textit{100.0} & \textit{99.80} & \textit{100.0} & \textit{99.40} & \textit{100.0} & \textit{99.95} & \textit{96.48} & \textit{98.73} & \textit{100.0} & \textit{85.00} & \textit{68.00} & \textit{95.94} \\

MiniCache & 29.80 & 7.00 & 0.00 & 2.80 & 0.00 & 0.00 & 0.90 & 0.80 & 4.10 & 50.30 & 36.20 & 41.60 & 32.60 & 15.85  \\

CommonKV &  100.00 & 99.80 & 100.00 & 100.00 & 99.80 & 99.40 & 99.90 & 99.95 & 90.28 & 94.73 & 100.00 & 82.80 & 64.60 & 94.71 \\
\hdashline

\methodnoavg{} & 84.80 & 81.80 & 24.00 & 86.40 & 79.00 & 5.00 & 84.50 & 50.90 & 15.82 & 32.33 & 57.68 & 76.40 & 60.40 & 56.85 \\
\method{} & 100.00 & 99.60 & 100.00 & 99.60 & 100.00 & 99.20 & 99.70 & 96.45 & 83.26 & 97.80 & 99.68 & 82.60 & 65.20 & 94.08 \\

\rowcolor{gray!15}[0pt][0pt] \multicolumn{15}{c}{\textit{Qwen3-8B }} \\

\textit{Full KV} & \textit{100.0} & \textit{100.0} & \textit{99.80} & \textit{99.80} & \textit{100.0} & \textit{99.80} & \textit{99.85} & \textit{99.55} & \textit{83.76} & \textit{93.80} & \textit{100.0} & \textit{75.40} & \textit{59.00} & \textit{93.14} \\

MiniCache & 0.60 & 0.00 & 0.00 & 0.00 & 0.00 & 0.00 & 0.00 & 0.00 & 0.50 & 15.30 & 0.00 & 6.60 & 8.00 & 2.39 \\

CommonKV & 100.00 & 100.00 & 98.60 & 98.40 & 98.40 & 74.20 & 99.60 & 97.00 & 57.78 & 93.33 & 99.04 & 66.20 & 54.40 & \textbf{87.46} \\
\hdashline

\methodnoavg{} & 53.20 & 25.60 & 3.40 & 24.40 & 0.80 & 0.00 & 5.85 & 5.55 & 4.02 & 0.13 & 8.76 & 58.80 & 44.80 & 18.10 \\

\method{} & 99.80 & 100.00 & 98.20 & 99.00 & 99.40 & 41.20 & 99.65 & 99.30 & 57.58 & 86.47 & 99.08 & 72.20 & 55.40 & \underline{85.18} \\

\bottomrule
\end{tabular}
}
\end{table*}

\begin{table}[h]
\caption{Hyperparameter settings for CommonKV on each model. Group size is fixed at 4 for all experiments. We select the number of groups to merge while keeping the rank around 0.7 as per the original paper, but allow the rank to vary to accommodate the exact compression ratio.}
\centering
\begin{tabular}{clcccc}
\toprule
Values Ratio& Model & Layers &  Merged Group Fraction & Rank \\
\midrule
\multirow{3}{*}{50\%} & Llama-3.1-8B-Instruct & 32 & 5/8 & 0.7008 \\
& Qwen3-8B & 36 & 6/9 & 0.7440 \\
& Mistral-Small-3.1-24B-Inst. & 40 & 8/10 & 0.7411 \\ \cmidrule{2-6}
\multirow{3}{*}{75\%} & Llama-3.1-8B-Instruct & 32 & 6/8 & 0.7069 \\
& Qwen3-8B & 36 & 7/9 & 0.7416 \\
& Mistral-Small-3.1-24B-Inst. & 40 & 9/10 & 0.7565 \\ 
\bottomrule 
\end{tabular}
\label{tab:rank}
\end{table}

\section{CommonKV settings}
\label{app:commonkv_settings}

We report exact CommonKV group and rank ratio hyperparameter settings for the three models we test, as well as the two values compression ratios, found in \Cref{tab:rank}.
For all CommonKV experiments, the first 4 sink tokens and 128 recent tokens (non-sliding, as CommonKV compresses prefill  only) are not merged as per the proposed method. 
Finally, we adopt the paper's Fisher warmup settings of 2048 samples of length 1024 sequences of Wikitext2.

\section{Timing Experiments Details}
\label{app:timing}

All efficiency measurements are computed using a single NVIDIA A100-80GB GPU with PyTorch 2.5.1, CUDA 12.1, in float16. 
Attention is implemented using SDPA. We build the length 8192 prompts by sampling token IDs uniformly at random over the vocabulary and greedy decoding for 256 tokens. 

TPOT is the mean over the 255 decoding steps, because the first decoding step is computed from prefill. TTFT is the total time from the start of prefill until the sampling of the first token. We use 2 warm-up and 3 timed trials, and report the mean.

\end{document}